\documentclass[letterpaper, 10 pt, conference]{ieeeconf}  

\IEEEoverridecommandlockouts                              

\usepackage{amsmath} 
\usepackage{amssymb}  
\usepackage[ruled]{algorithm2e}
\usepackage{bm}
\usepackage{braket}
\usepackage[T1]{fontenc}

\let\labelindent\relax
\usepackage[inline]{enumitem}

\usepackage{float}
\usepackage{placeins}
\usepackage{hyperref}
\usepackage{graphicx}

\title{\LARGE \bf
    Dynamic Decentralized 3D Urban Coverage and Patrol with UAVs
}

\author{
    Wai Lun Leong\authorrefmark{1}\authorrefmark{2}, Jiawei Cao\authorrefmark{1}, and Rodney Swee Huat Teo\authorrefmark{1}
    \thanks{\protect\\ \authorrefmark{1}National University of Singapore, Singapore.\protect\\ \authorrefmark{2}Corresponding author. Email: {\tt\small tsllwl@nus.edu.sg}}%
}

\begin{document}

\maketitle
\thispagestyle{empty}
\pagestyle{empty}

\begin{abstract}

In the event of natural or man-made disasters in an urban environment, such as fires, floods, and earthquakes, a swarm of unmanned aerial vehicles (UAVs) can rapidly sweep and provide coverage to monitor the area of interest and locate survivors. We propose a modular framework and patrol strategy that enables a swarm of UAVs to perform cooperative and periodic coverage in such scenarios. Our approach first discretizes the area of interest into viewpoints connected via closed paths. UAVs are assigned to teams via task allocation to cooperatively patrol these closed paths. We propose a minimal, scalable, and robust patrol strategy where UAVs within a team move in a random direction along their assigned closed path and ``bounce'' off each other when they meet. Our simulation results show that such a minimal strategy can exhibit an emergent behaviour that provides periodic and complete coverage in a 3D urban environment.

\end{abstract}

\section{Introduction}
\label{sec:introduction}



In recent years, the urban setting has become a common backdrop for natural or man-made disasters. In such settings, the vertical dimension becomes an important consideration and the role of unmanned aerial vehicles or UAVs for search and rescue purposes have become increasingly prevalent~\cite{polka2017use}. A cluttered, low-rise and built-up urban neighbourhood can be cooperatively searched and continuously monitored using multiple UAVs to locate survivors who may have emerged from or become trapped in buildings, or are awaiting rescue in the aftermath of events such as fires, floods, or earthquakes.

In the 3D urban coverage and patrol problem, a team of UAV agents equipped with cameras is assigned to cover and patrol an urban environment that consists of buildings represented by right prisms. The facades and roof of each building within this environment must be covered with some image quality constraints to enable object detection, therefore, an agent cannot simply move until the entire facade is fully visible within its camera field of view (FOV). In addition, we also assume that there are not enough agents to provide continuous coverage over the environment and therefore the agents must move or patrol continuously and provide periodic coverage over each part of the facade.

In the context of this problem, we consider the use of multirotor UAV agents. Multirotor UAVs offer advantages over their fixed-wing counterparts such as vertical takeoff and landing, hovering capability, and ease of operation, making them suited for tasks requiring precise maneuvering in confined or dynamic urban environments. Their ability to operate at lower altitudes and excel in close proximity missions further enhances their suitability for applications such as inspection, surveillance, and search and rescue.

In the following, we examine related works that enable multi-robot systems to cooperatively solve the coverage and patrol problem in both 2D and 3D domains, and note that most of such works are in the 2D domain. We suggest that both cases are relevant to our problem as the 2D and 3D coverage and patrol problem can often be reduced to a 1D closed path patrol problem.

Freda et. al. described a distributed 3D multi-robot patrolling and coordination strategy~\cite{freda20193d}. Their two-level coordination strategy considers both topological and metric aspects of the problem. The topological strategy enables each agent to select and deconflict the next node to visit to minimize the average global idleness, while the metric strategy performs path planning and deconfliction. Their proposed approach is applied to unmanned ground vehicles patrolling a 2D manifold in 3D space but does not consider the complete coverage of buildings.

Piciarelli and Foresti proposed an algorithm for cooperative coverage and patrol using reinforcement learning applied to a swarm of drones in an environment where the different parts of the environment have different coverage priorities~\cite{piciarelli2020drone}. The problem is modeled as a Markov
Decision Process (MDP) where agents can take various actions such as moving or zooming the camera, resulting in a reward. Reinforcement learning is used to train a Deep Q-Network (DQN) model to achieve prioritized drone coverage. However, their approach does not consider topological altitudes of the environment or buildings.

Cooperative perimeter surveillance strategies have been proposed for problems such as monitoring of forest fires~\cite{casbeer2006cooperative, sujit2007cooperative, ghamry2016cooperative} and perimeter defense~\cite{matlock2009cooperative}. These solutions assume that a perimeter can be formed over the area of interest such that multiple agents can cooperatively partition the path and perform surveillance using a frequency-based approach. Such strategies have inherent properties of robustness and fault tolerance as agents meet at some interval for message and information exchange to improve situation awareness for coordination, decision making, and reporting.

Various strategies and theoretical optimal solutions have been proposed for the multi-agent boundary and circular patrol problems by \cite{czyzowicz2011boundary} and \cite{kawamura2015simple}.

Kingston et. al. proposed a method for decentralized perimeter surveillance using agents modeled as point masses that move at uniform constant velocity, which provably converges to an optimal behaviour in finite time and accounts for communications constraints~\cite{kingston2008decentralized}. Each agent moves along the segment of the linear path it is responsible for and reverses direction upon reaching the endpoints. When an agent encounters another agent, it re-partitions the path given the perimeter length and the number of agents relative to its left and right (determined via message exchange of coordination variables), and moves to patrol its new perimeter segment. When the algorithm converges and agents are synchronized in time and space, the oscillation of each agent within its perimeter segment result in surveillance along the perimeter as well as periodic communications connectivity and information exchange. The authors suggest that this can be visualized as beads sliding along a wire.

A similar frequency-based approach is proposed by Acevedo et. al in \cite{acevedo2013cooperative1} by partitioning the problem between homogeneous agents for 2D area coverage. The authors applied the same method to cooperatively patrol a closed path within a 2D area with heterogeneous agents in \cite{acevedo2013cooperative2}, which was improved for faster convergence in \cite{acevedo2013distributed}.

We propose a decentralized and modular framework and patrol strategy for the multi-agent 3D urban coverage and patrol problem. There are two main problems to be solved: coverage path planning, where we partition a 3D area of interest and generate a minimal path within each partition, and the patrol strategy, where decentralized agents assigned to a particular path coordinate to partition the path and patrol to provide coverage.

Our method works by breaking the problem down into \emph{viewpoint generation}, \emph{task generation}, \emph{task allocation}, and \emph{patrol strategy}. We discretize and generate viewpoints over buildings in an urban environment, connect the viewpoints belonging to each building to form a closed path, perform task allocation of multiple agents to each closed path, and finally partition and patrol each closed path using a minimal approach where agents ``bounce'' off each other along the path. Our modular approach to the problem allows each part of the problem to be solved and improved easily.

In contrast with \cite{kingston2008decentralized, acevedo2013cooperative2, acevedo2013distributed}, which aim to optimally partition a continuous 2D perimeter for surveillance, our approach conducts surveillance over discrete viewpoints along 3D closed paths. This is achieved through emergent behaviour resulting from the exchange of local information, without relying on global information and consensus. In addition, our approach directly addresses the coverage path planning problem in a 3D environment and our patrol strategy minimizes computation and communication requirements, making it ideal for resource-constrained platforms operating in urban settings where communication signals may be sporadic or limited in range.


\section{Problem Formulation}
\label{sec:problem_formulation}

We define an a priori 3D environment by a set of buildings $\bm{B} = \{ \bm{B}_1, \bm{B}_2, \bm{B}_3, \dots \}$. Each building $\bm{B}_j$ is an extruded 2D polygon or right prism described by a tuple consisting of a 2D polygon $\bm{P}_j$ with $n$ 2D points, representing its 2D footprint on the ground level and height $h_j$:

\begin{equation}
\begin{aligned}
    \bm{P}_j &= [(x_1,y_1), (x_2,y_2), \dots, (x_n,y_n)] \\
    \bm{B}_j &= \set{ \bm{P}_j, h_j } \\
    &\quad\quad\quad\quad\lvert \bm{P}_j \rvert \geq 3, \quad h_j > 0
\end{aligned}
\end{equation}

Given $\bm{P}_j$ and $h_j$, we can generate the 2D polygon representing the roof surface in 3D coordinates $\bm{R}_j$:

\begin{equation}
\begin{aligned}
    \bm{R}_j = [(x_1,y_1,h_j), (x_2,y_2,h_j), \dots, (x_n,y_n,h_j)]
\end{aligned}
\end{equation}

For the same $\bm{P}_j$ and $h_j$, we generate $n$ vertical 2D surfaces in 3D coordinates representing the facades of the building:

\begin{equation}
\begin{aligned}
    \bm{F}_{j_1} &= [ (x1,y1,0), (x2,y2,0), (x2,y2,h_j), (x1,y1,h_j)] \\
    \bm{F}_{j_2} &= [ (x2,y2,0), (x3,y3,0), (x3,y3,h_j), (x2,y2,h_j)] \\
    &\dots \\
    \bm{F}_{j_n} &= [ (x_n,y_n,0), (x1,y1,0), (x1,y1,h_j), (x_n,y_n,h_j)] \\
    \bm{F}_j &= [ \bm{F}_{j_1}, \bm{F}_{j_2}, \dots, \bm{F}_{j_n} ] \\
\end{aligned}
\end{equation}

The building $\bm{B}_j$ can then also be represented by the set of 3D surfaces $\bm{S}_j$ derived from its original definition:

\begin{equation}
    \bm{S}_j = \bm{F}_j \cup \bm{R}_j = [ \bm{F}_{j_1}, \bm{F}_{j_2}, \dots, \bm{F}_{j_n}, \bm{R}_j ]
\end{equation}

For the scope of this work, we define surfaces as the vertical facades and roofs of buildings, but this method can be applied to any general 3D urban scenario as long as its features can be decomposed into 3D surfaces, such as roads, points and areas of interest at ground level, etc.

We assume that our UAV agents are equipped with a gimbal mount that allows the camera to tilt between forwards and downwards with respect to the UAV's body frame. This will allow the agent to view facade surfaces by facing the facade and pitching the camera forwards, as well as roof surfaces by flying above the roof and pitching the camera downwards. Given the camera FOV spanning $c_w$ degrees in width and $c_h$ degrees in height, we select $c_d$ such that an optimal resolution of the surface is obtained for object detection, the width $f_w$, height $f_w$, and area $a_f$ of the rectangular polygon $p_v$ on the surface that can be instantaneously observed by the camera is:

\begin{equation}
\begin{aligned}
    f_w &= 2 \times c_d \times \tan\left(\frac{c_w}{2}\right) \\
    f_h &= 2 \times c_d \times \tan\left(\frac{c_h}{2}\right) \\
    a_f & = f_w \times f_h
\end{aligned}
\end{equation}

Since we do not tilt the camera in directions other than forwards and downwards with respect to the UAV, the above is sufficient to represent the camera FOV. In this work, we assume a homogeneous camera capability for all agents and for heterogeneous cases, it is sufficient to assume the minimum case.

Given a surface $s \in \bm{S}_j$ and its area $a_s$, we discretize it into a set of associated viewpoints $\bm{V}_s$, such that:

\begin{equation}
\begin{aligned}
   \lvert \bm{V}_s \rvert = \Bigl\lceil \frac{a_s}{a_f} \Bigr\rceil
\end{aligned}
\end{equation}

We define $\bm{V}_j$ as the set of all viewpoints belonging to $\bm{S}_j$:

\begin{equation}
\begin{aligned}
    \bm{V}_j = \bigcup \bm{V}_s \quad \forall s \in \bm{S}_j
\end{aligned}
\end{equation}

Since each set of surfaces $\bm{S}_j$ is associated with a building $\bm{B}_j$, the set of all viewpoints in the scenario $\bm{V}$ is:

\begin{equation}
\begin{aligned}
    \bm{V} = \bigcup \bm{V}_j \quad \forall j \in  \bm{B}
\end{aligned}
\end{equation}

Similar to \cite{machado2002multi}, we define the \emph{idleness} $i_v$ as the period between two consecutive visits to a viewpoint in $v \in \bm{V}$ and the set of $i_v$, $\forall v \in \bm{V}$ as $\bm{I}$. Our objective is to minimize the maximum $i_v \in \bm{I}$.

We propose to apply our approach to a swarm of UAV agents, and therefore subject it to the following attributes, requirements and constraints similar to our prior work in \cite{leong2021intelligent}:

\begin{itemize}
    \item \emph{Decentralization}: Agents behave as peers without special roles and make decisions based on local information, increasing flexibility, reliability, and robustness.
    \item \emph{Heterogeneous}: Problems are modeled as heterogeneous tasks (e.g. ``3D coverage and patrol'', ``target tracking'', ``monitor point of interest''), to be allocated to a set of heterogeneous agents (e.g. fixed wing UAVs, multirotor UAVs, ground robots).
    \item \emph{Scalability}: The addition of agents and tasks should not have a significant impact on performance.
    \item \emph{Adaptability}: Dynamic and online reconfiguration due to changes in available agents and mission tasks.
    \item \emph{Reliability}: The system is fault tolerant and robust, automatically reconfiguring if some drones experience partial or total failure, detected by loss of communications.
    \item \emph{Communication}: Local interactions such as broadcasting of messages through radios with limited range and bandwidth.
    \item \emph{Convergence}: Stable, near-optimal solutions are obtained in a finite and deterministic time horizon.
\end{itemize}

\section{3D Urban Coverage and Patrol}
\label{sec:3d_urban_cp}

We divide the problem into 4 parts to be solved:

\begin{itemize}
    \item \emph{Viewpoint Generation}: Generate viewpoints from 3D surfaces that represent buildings
    \item \emph{Task Generation}: Generate tasks consisting of closed paths through sets of viewpoints, where each set belongs to a single building
    \item \emph{Task Allocation}: Allocate agents to tasks
    \item \emph{Patrol Strategy}: Agents allocated to a task partition and patrol along its associated closed path
\end{itemize}

With the above, we can create a modular solution that can be easily improved or modified to optimise for different objective functions by changing the solution for the relevant part.

In the following, we will describe in detail each of the components in our proposed solution.

\subsection{Viewpoint Generation}
\label{subsec:viewpoint}

Given a surface $\bm{S}_j$ such as a building facade or roof, we discretized it in the previous section into a set of viewpoints $\bm{V}_j$. Each viewpoint $v \in \bm{V}_j$ provides complete coverage of a rectangular polygon in $p_v$ in 3D space on the surface of some surface $s \in \bm{S}_j$, $f_w \times f_h$ in size, corresponding to agent camera's FOV at a distance of $c_d$. For facades, the edges of $p_v$ are parallel and perpendicular to the ground plane; for roofs, they are parallel and perpendicular to the principal axis of the roof surface $\bm{R}_j$.

Each viewpoint $v$ is defined by $(x_v, y_v, z_v, b_v, t_v)$, where $x_v, y_v, z_v$ is the 3D position of the agent that allows it to look at $p_v$, $b_v$ is the compass bearing representing the orientation of the agent in the world frame, and $t_v$ is the tilt of the camera with respect to the body frame of the agent. $x_v, y_v, z_v$ can be obtained by a tangential projection of $c_d$ from the centroid of each $p_v$. $b_v$ and $t_v$ are set such that the agent will yaw and the camera will tilt towards the centroid of $p_v$ and the camera's FOV $p_f$ aligns with $p_v$.

\begin{figure}[htbp]
    \centering
    \includegraphics[width=0.75\columnwidth]{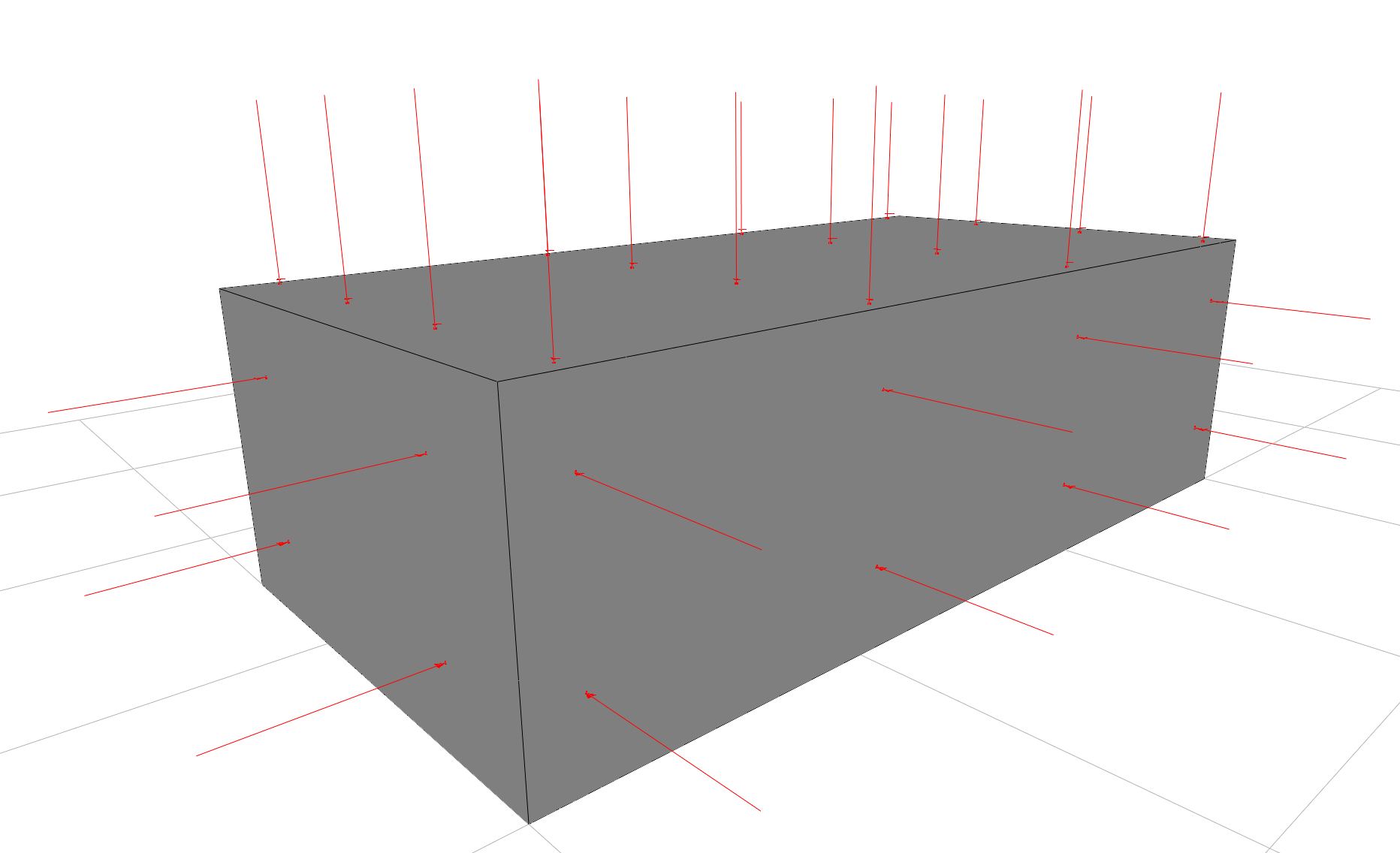}
    \caption{Example of viewpoints generated for a single building.}
    \label{fig:1bld_viewpoints}
\end{figure}

We described how we generate viewpoints for an urban environment consisting of buildings modeled as right prisms. An example of the generated viewpoints for a single building is shown in Fig.~\ref{fig:1bld_viewpoints}.

\subsection{Task Generation}

From building $\bm{B}_j$, we obtained a set of associated viewpoints $\bm{V}_j$. The goal of task generation is to create a closed path $\bm{H}_j$ for each $\bm{B}_j$ that passes through every viewpoint in $\bm{V}_j$ to create a task $j$. When one or more agents moves or patrols along $\bm{H}_j$, visiting every $v \in \bm{V}_j$, complete coverage over the building will be achieved.

Each task $j$ consists of an associated closed path $\bm{H}_j$, and its \emph{capacity} $c_j$ ($j = \{ \bm{H}_j, c_j \}$), which determines how many agents will be assigned to $j$. We compute the capacity given the number of available agents and the length of each $\bm{H}_j$ to give an even distribution, but other schemes can be used to give a better result for a different objective function, e.g. to prioritize the minimization of idleness values for a particular set of viewpoints by assigning more agents.

\begin{figure}[htbp]
    \centering
    \includegraphics[width=0.75\columnwidth]{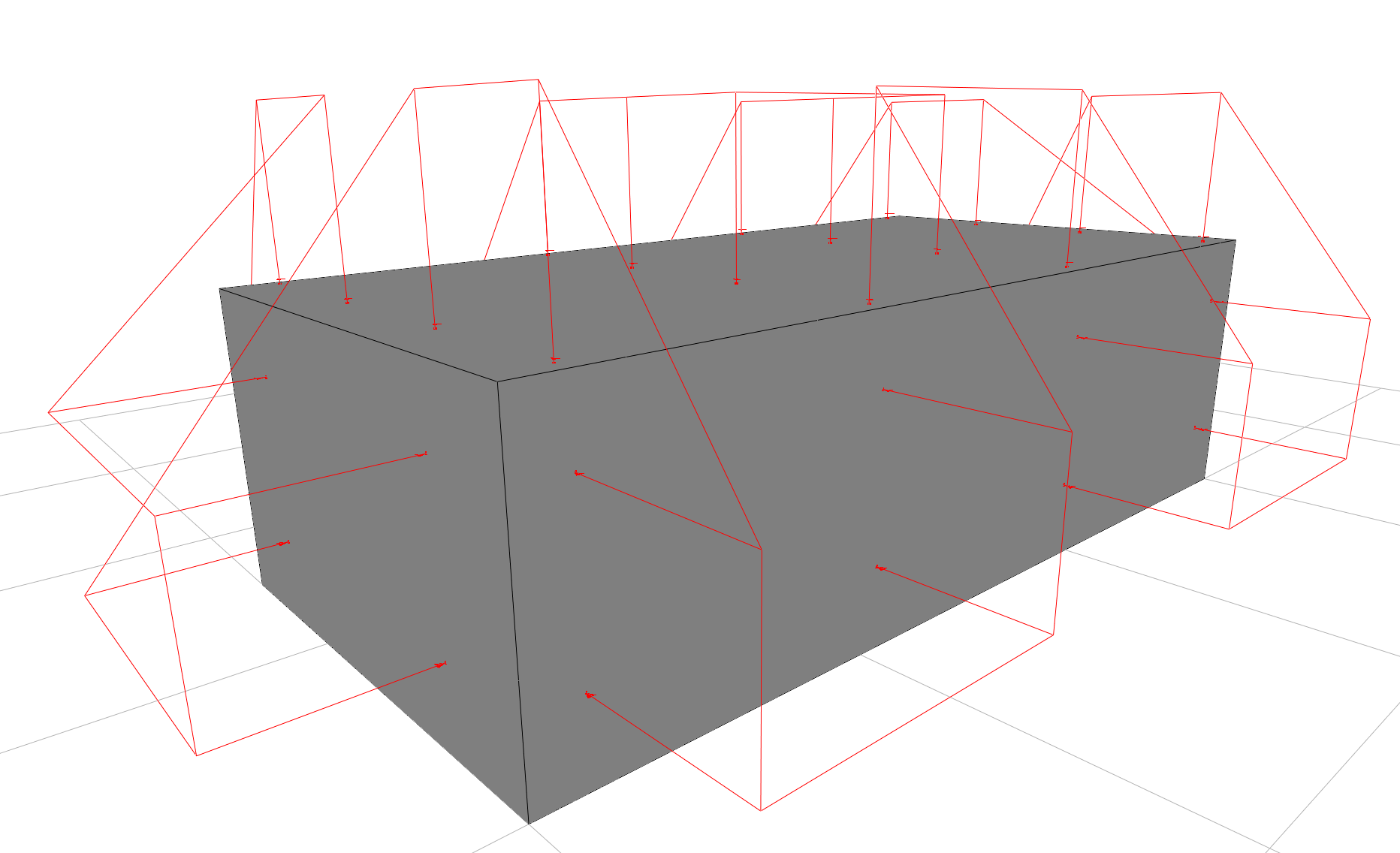}
    \caption{Example of viewpoints with closed path.}
    \label{fig:1bld_viewpoints_tsp}
\end{figure}

Since $\bm{H}_j$ is a closed, ordered list of viewpoints in $\bm{V}_j$, the generation of such a closed path while minimizing its length is essentially a 3D version of the Traveling Salesman Problem (TSP). Since the TSP is NP-hard~\cite{Karp1972}, it is not practical to obtain an optimal solution for larger problem sizes and therefore we use the Christofides algorithm, a heuristic that guarantees the result to be within 1.5 times of the optimal solution~\cite{christofides1976worst}. An example of the resulting closed path can be seen in Fig~\ref{fig:1bld_viewpoints_tsp}.

A non-trivial point for the implementation of the above to note is that the distance function between two viewpoints should not simply consider the 3D Euclidean distance, but also the presence of obstacles, including the building itself. Viewpoints opposite the same building should consider the distance based on a path that avoids the building itself, which can be done using 3D visibility graphs~\cite{huang2019computationally}.

Although we proposed that each building $\bm{B}_j$ is associated with a single task $j$, this is a design decision rather than a constraint and any given set or union of sets of viewpoints can be used to generate $\bm{H}_j$. For example, $j$ may have a $\bm{H}_j$ that consists of viewpoints from 2 adjacent buildings, or a building and adjacent roads surrounding the building. It is also possible to generate a single closed path for every viewpoint for a given scenario, resulting in a single task. The advantage of dividing the scenario into multiple tasks in the manner that we have described above is that smaller dynamic changes such as the loss of an agent for a particular task will cause changes only to agents allocated to the task and will not result in a disturbance for other tasks.

From the above, we obtained a set of tasks $\bm{T} = \set{\bm{T}_1, \bm{T}_2, \bm{T}_3, \dots, \bm{T}_n}$ corresponding to the set of buildings $\bm{B} = \set{\bm{B}_1, \bm{B}_2, \bm{B}_3, \dots, \bm{B}_n}$. This step only needs to be performed once for every $\bm{B}$ and requires more computing power, but it can be performed offline, or at the ground control station (GCS) which we assume to have sufficient computing capability.

The step of \emph{task dissemination} so that all agents have the same copy of $\bm{T}$ for task allocation is efficiently achieved through our approach based on distributed blockchains, but is out of the scope of this work.

\subsection{Task Allocation}

We allocate the set of agents $\bm{A}$ to the set of tasks $\bm{T}$ and assume that $\lvert \bm{A} \rvert \geq \lvert \bm{T} \rvert$. Many of our desired attributes for a solution that aligns to the swarm paradigm are handled in this step. In particular, we desire a task allocation approach that is decentralized, and can handle heterogeneous tasks and agents to achieve scalability, adaptability and fault tolerance.

The Consensus-Based Bundle Approach (CBBA)~\cite{choi2009consensus} is an auction-based approach for decentralized task allocation that we have successfully implemented and demonstrated in our prior work (\cite{leong2021intelligent, leong2022pheromone}). Agents iterate between two phases: a bundle building phase where each agent uses score functions to greedily generate bids for an ordered bundle of tasks to execute, and a consensus phase where agents communicate with neighbours to deconflict tasks. Each agent acts as a decentralized relay, consolidating and broadcasting bids for tasks to neighbours, resulting in global consensus even without a fully connected communications graph. In addition, score functions to generate bids allow us to consider constraints to heterogeneous tasks such as agent capabilities, endurance, task-agent type matching, task priorities, etc. To allow multiple agents to be assigned to a single task, we implement the extension Consensus-Based Grouping Algorithm (CBGA)~\cite{hunt2014consensus}. We suggest that the readers refer to the above papers for further details. We set the bundle size $L_t = 1$ as we do not want agents to be moving from task to task (building to building).

Similar to \cite{choi2009consensus}, we define $x_{ij} \in \{0, 1\}$ as a binary decision variable to indicate if agent $i$ is assigned to task $j$ and therefore:

\begin{equation}
\begin{aligned}
    &\sum_{j=1}^{\lvert \bm{T} \rvert } x_{ij} \leq 1 \quad \forall i \in \bm{A} \\
    &\sum_{i=1}^{\lvert \bm{A} \rvert } x_{ij} \leq c_j \quad \forall j \in \bm{T}   
\end{aligned}
\end{equation}

Our task allocation implementation runs as an online and anytime algorithm, giving us the qualities of scalability, adaptability and fault tolerance.  For the score function $s_{ij}$ for agent $i$ with respect to task $j$ , we consider the squared distance $d_{ij}$ from $i$ $(x_i, y_i, z_i)$ to the centroid of the building $\bm{B}_j$ associated with $j$ $(x_j, y_j, z_j)$:

\begin{equation}
\begin{aligned}
    d_{ij} = \max \{(x_i - x_j)^2 + &(y_i - y_j)^2 + (z_i - z_j)^2, 1\} \\
    d_{ij} &\geq 1
\end{aligned}
\end{equation}

We use square distance since the score used for bids are relative and not a representation of distance. Next, we normalize the above to obtain a score which will be used to select and bid for tasks by $i$:

\begin{equation}
\label{eq:score}
\begin{aligned}
    s_{ij} &= \frac{1}{d_{ij}} \\
    s_{ij} &> 0
\end{aligned}
\end{equation}

A useful formulation to improve (\ref{eq:score}) is to add a non-negative integer $p_j$ to represent the priority of task $j$ ($j = \{ \bm{H}_j, c_j, p_j \}$) to override the distance factor and ensure that higher priority tasks associated with buildings of higher importance are allocated with agents first:

\begin{equation}
\begin{aligned}
    s_{ij} &= p_j + \frac{1}{d_{ij}} \\
    s_{ij} &> 0, \quad p_j \in \mathbb{Z}^{\geq 0}
\end{aligned}
\end{equation}

It is important to normalize the score with respect to other task types in a heterogeneous task scenario. For example, during the execution of task $j$, an object of interest is detected in the polygon $p_v$ covered by viewpoint $v$. A tracking task can be generated at a higher priority than $j$, so that an agent can be reassigned through task allocation to continuously track the object of interest, while the remaining agents reconfigure to handle the original set of tasks.

The task allocation process results in the set of agents $\bm{A}$ being assigned to the set of tasks $\bm{T}$. This assignment is performed in a decentralized manner using the greedy auction-based method CBBA and CBGA, deconflicting the results such that each agent is assigned to not more than 1 task and each task is assigned a number of agents not more than its capacity constraint.

\subsection{Patrol Strategy}

Given a task consisting of a closed path through multiple viewpoints with one or more agents assigned to it, we require a patrol strategy such that all viewpoints will be periodically traversed and the path is partitioned among the assigned agents. The set of agents can assigned to the task can change during the mission due to reinforcements or losses and therefore the algorithm must be decentralized, adaptable, reliable and scalable. Each agent is constrained by size, weight and power (SWaP) limitations, so the computational and communications complexity of the algorithm should not increase as the number of agents increase.

In addition, agents have limited broadcast communications capabilities, and may not be able to communicate due line-of-sight (LOS) constraints in an urban environment. It is important that agents are able to exchange data to enable coordination, commands from the GCS such as new tasks, and be able to report relevant information such as the detection of objects of interest back to the GCS. In many practical cases, when continuous connectivity is not possible, periodic connectivity can still achieve the desired results. This means that an agent does not remain in continuous communication with other agents all the time, but meets and exchanges information with other agents periodically (periodic connectivity).

In the previous sections, we have reformulated the original problem of 3D urban coverage and patrol into a closed perimeter surveillance problem by discretizing the area of interest into viewpoints and generating a 3D closed path along the viewpoints. In this section, we present our solution for the above while aiming to satisfy the requirements with a \emph{minimum} amount of computation and communications to enable scalability.

The patrol strategy to be developed is with respect to a particular task $j$ and its associated patrol path $\bm{H}_j$. The task of agents $\bm{A}_j$ assigned to $j$ is to move along $\bm{H}_j$, stop at every viewpoint to perform object detection, and continue onto the next viewpoint. We attempt to minimize the idleness $i_v$, representing the idleness of a viewpoint $v$, $\forall v \in \bm{H}_j$.

We propose a \emph{minimal} strategy where each agent in $\bm{A}_j$ assigned to $j$ join $\bm{H}_j$ via its respective nearest viewpoint, and move in a random direction along $\bm{H}_j$. Since $\bm{H}_j$ is a 3D path in an urban environment, we do not want an increased risk of collisions caused by agents moving past each other. Our desired behaviour is to have agents moving in opposite directions along $\bm{H}_j$ meet, exchange messages, and ``bounce'' off each other.

Given that $\bm{H}_j$ is a closed path, for any $\lvert \bm{A}_j \rvert \geq 2$, agents moving in different directions or speeds will eventually meet each other along $\bm{H}_j$ and exchange messages, resulting in oscillation, path partitioning, and periodic connectivity regardless of the length of $\bm{H}_j$ or cardinality of $\bm{A}_j$.

Our proposed approach is also scalable and robust against changes in the number of agents assigned to the task, as agents only interact with neighbouring agents within communications range, making decisions based on local information without global consensus.

For example, if agents $i$ and $k$ would have met at some point in time and agent $k$ was removed, agent $i$ simply continues travelling in the same direction, extending its patrol path and covering the viewpoints that $k$ would have covered. Suppose another agent was inserted between agents $i$ and $k$, it would oscillate between both agents, reducing the number of viewpoints they would each need to cover and reducing the overall idleness.

Each agent $i$ stores and exchanges the following information vector with its neighbours:

\begin{equation}
\begin{aligned}
    \bm{M}_i &= [x_i, y_i, z_i, c_i, l_i, d_i] \\
    1 \leq c_i \leq \lvert \bm{H}_j \rvert, &\quad 1 \leq l_i \leq \lvert \bm{H}_j \rvert, \quad d_i \in \{ -1, 1 \}
\end{aligned}
\end{equation}

where $x_i, y_i, z_i$ is the 3D coordinates of $i$, $c_i$ is the index of the current viewpoint that the agent is servicing, $l_i$ is the index of the last viewpoint the agent has serviced, and $d_i$ represents the counter-clockwise and clockwise direction of the agent along $\bm{H}_j$.

We elaborate on the rules that enable agents to ``bounce'' off each other when servicing discrete viewpoints below and in Algorithm~\ref{alg:patrol}

\begin{figure}[htbp]
    \centering
    \includegraphics[width=0.75\columnwidth]{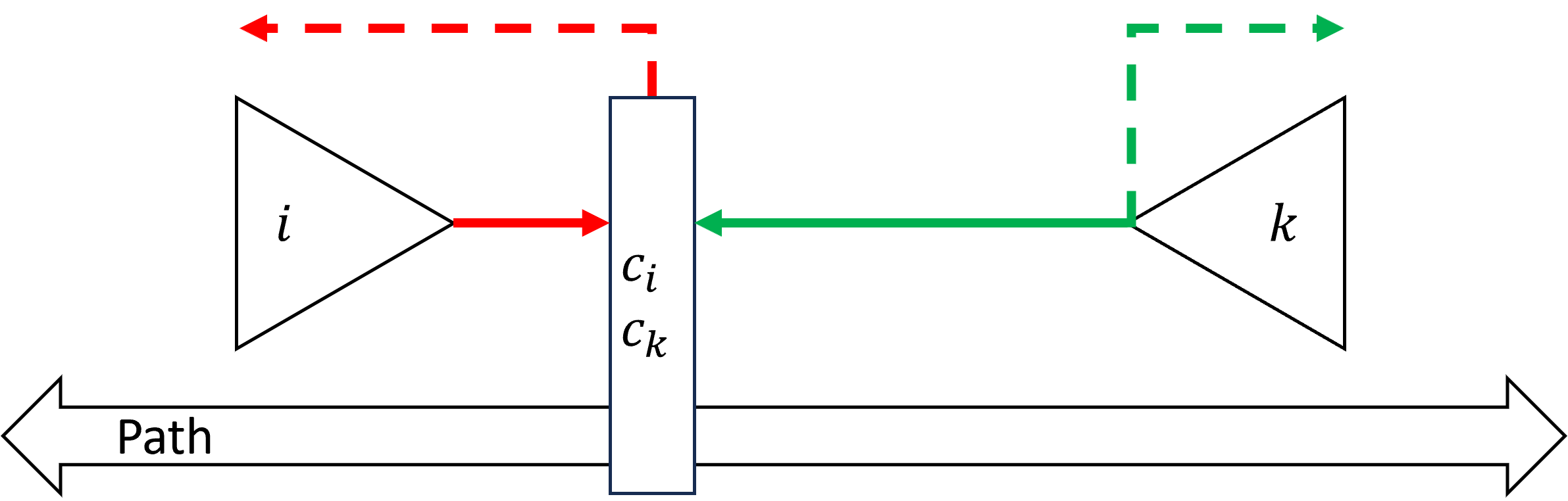}
    \caption{Agents $i$ and $k$, moving in opposite directions, have both selected the same viewpoint $v = c_i = c_k$, but since agent $i$ is nearer, it continues to service $v$ and changes direction after that, while agent $k$ changes direction immediately.}
    \label{fig:patrol_case1}
\end{figure}

Suppose two agents, $i$ and $k$ move into communication range with each other and exchange messages. If $c_i = c_k$, the agents are servicing the same viewpoint. To handle this conflict, we define the conflicting viewpoint $v = \bm{H}_j[c_i] = \bm{H}_j[c_k]$ and compare the distance of $i$ and $k$ with respect to $v$. The nearer agent continues to service $v$ and the further agent reverses direction immediately. For the nearer agent, two cases are possible. If both agents are moving in opposite directions ($d_i \neq d_k$) towards $v$, the nearer agent should reverse direction \emph{after} servicing $v$, as illustrated in Fig~\ref{fig:patrol_case1}.

\begin{figure}[htbp]
    \centering
    \includegraphics[width=0.75\columnwidth]{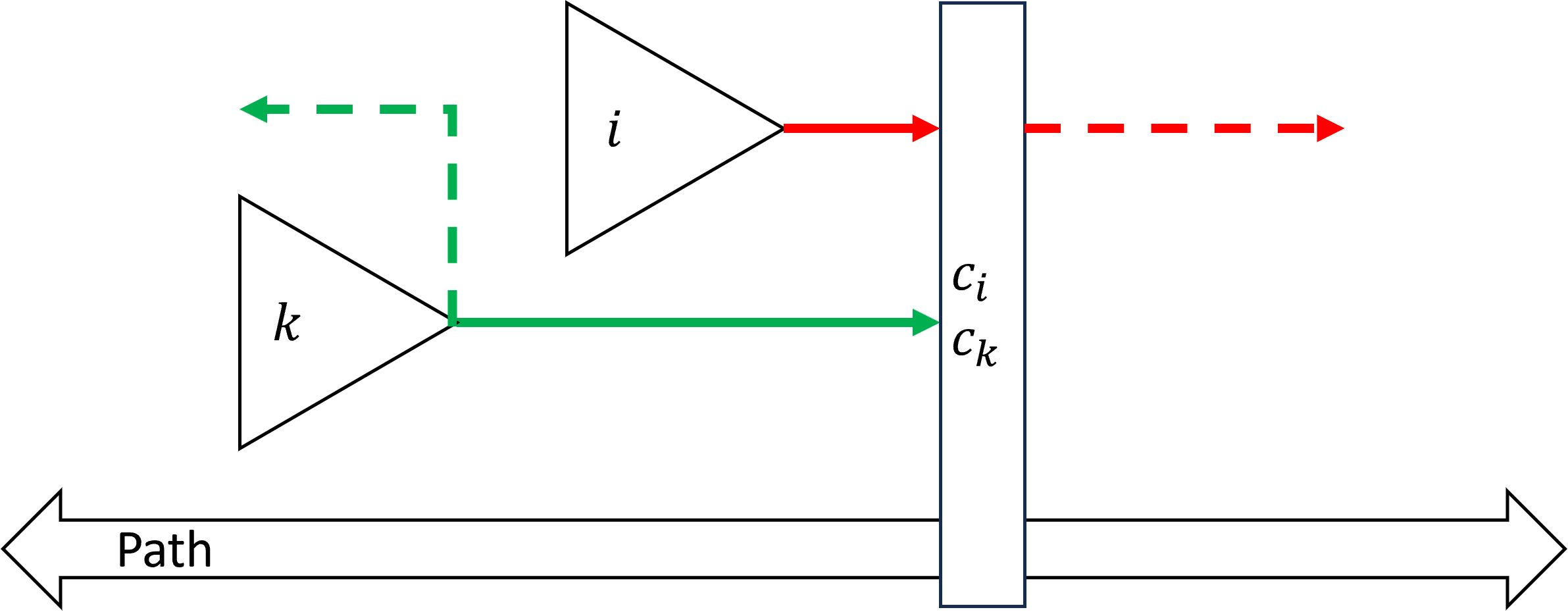}
    \caption{Agents $i$ and $k$, moving in the same direction, have both selected the same viewpoint $v = c_i = c_k$, but since agent $i$ is nearer, it continues to service $v$ and move in the same direction while agent $k$ changes direction immediately.}
    \label{fig:patrol_case2}
\end{figure}

In our next case, if both agents are moving in the same direction ($d_i = d_k)$, the nearer agent should continue in the same direction as only the further agent needs to change direction as shown in Fig~\ref{fig:patrol_case2}. In both cases, the desired behaviour is for $i$ and $k$ to end up moving in opposite directions.

\begin{figure}[htbp]
    \centering
    \includegraphics[width=0.75\columnwidth]{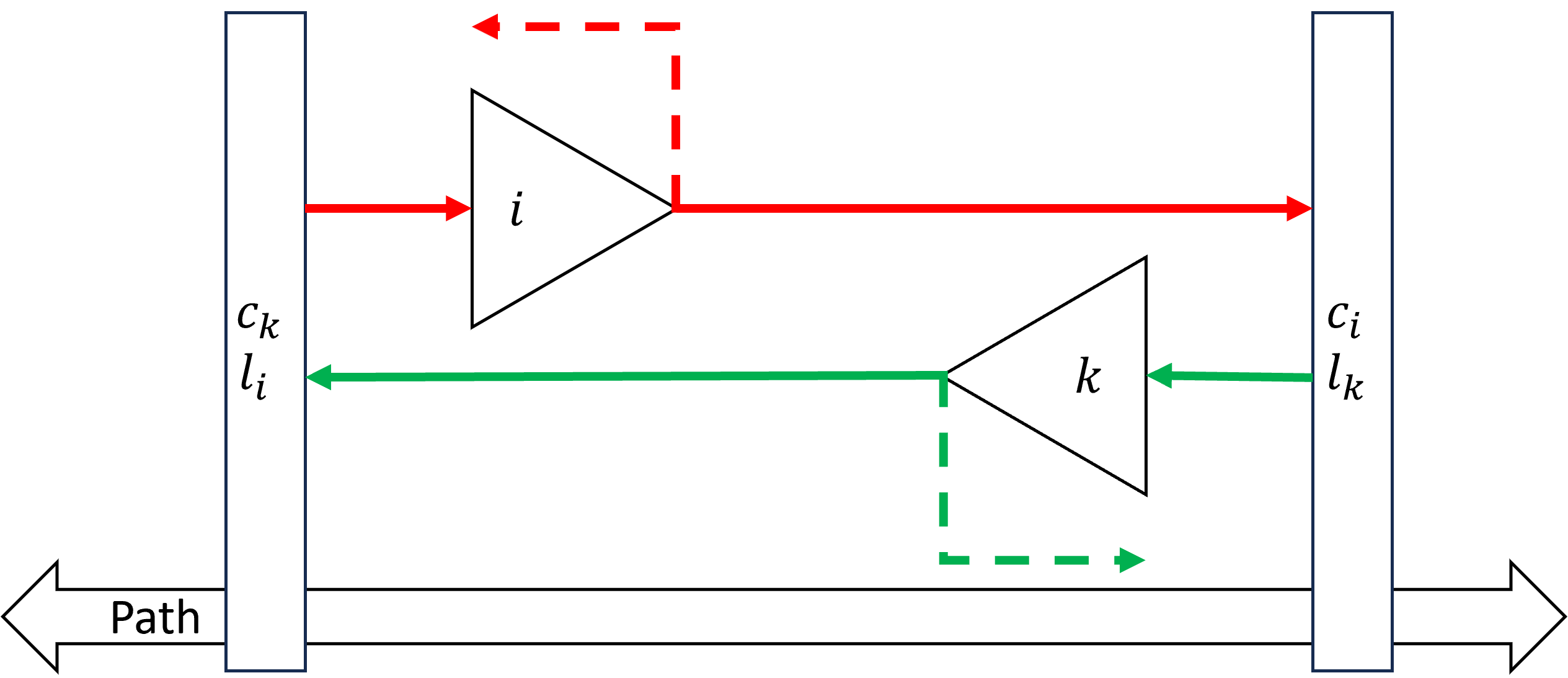}
    \caption{Agents $i$ and $k$, moving in the opposite directions and having selected each other's last viewpoints, have just come into communications range. Both agents reverse direction immediately.}
    \label{fig:patrol_case3}
\end{figure}

We examine a final case where $i$ and $k$ move into communication range with each other and exchange messages, but this time, $c_i = l_k \; \land \; c_k = l_i$. This means that the agents are moving towards each other's last viewpoint and also implies that the agents are moving in opposite directions along $\bm{H}_j$ and will pass each other. In this case, both agents reverse direction immediately as shown in Fig~\ref{fig:patrol_case3}.

\begin{algorithm}
    \caption{Decentralized Patrol Algorithm from perspective of agent $i$, periodically receiving and processing a set of messages $\bm{M}$}
    \label{alg:patrol}
    
    \For{each $\bm{M}_k \in M$}
    {
        \If{$c_i = c_k$}
        {
            \If{$\text{distance}(i, c_i) < \text{distance}(k, c_i)$}
            {
                $d_i = d_i \times -1$
            }
            \ElseIf{$d_i \neq d_k$}
            {
                scheduleChangeDirection = true
            }
            break
        }
        \ElseIf{$c_i = l_k \; \land \; c_k = l_i$}
        {
            $d_i = d_i \times -1$ \\
            break
        }
    }
    
    $v = \bm{H}_j[c_i]$ \\
    
    \If{$(x_i, y_i, z_i, b_i, t_i) \approx (x_v , y_v , z_v, b_v, t_v)$}
    {
        Perform detection
    }
    \Else{
        Move agent and actuate camera to $v$
    }

    \If{detection complete}
    {
        $l_i = c_i$ \\

        \If{scheduleChangeDirection}
        {
            $d_i = d_i \times -1$ \\
            scheduleChangeDirection = false
        }

        $c_i = ((c_i - 1 + d_i + \lvert \bm{H}_j \rvert) \mod \lvert \bm{H}_j \rvert) + 1$
    }
\end{algorithm}

The scalability of our patrol strategy with regards to computation and communication is due to the fact that each agent only considers information from its immediate neighbourhood and does not attempt to achieve global consensus or consider some shared global state. We note that this advantage comes at the cost of being unable to obtain an optimal solution where the closed path is partitioned equally among all agents. An implication is that the revisit interval for a viewpoint is non-deterministic in most cases, which may be an advantage when patrolling in a hostile environment.

In addition, our solution is fault-tolerant and enables periodic communication as long as the task is serviced by $\lvert \bm{A}_j \rvert \geq 2$ agents moving in opposite directions or different speeds.

\section{Simulation Results and Discussion}
\label{sec:simulation_results}

\begin{figure}[htbp]
    \centering
    \includegraphics[width=0.75\columnwidth]{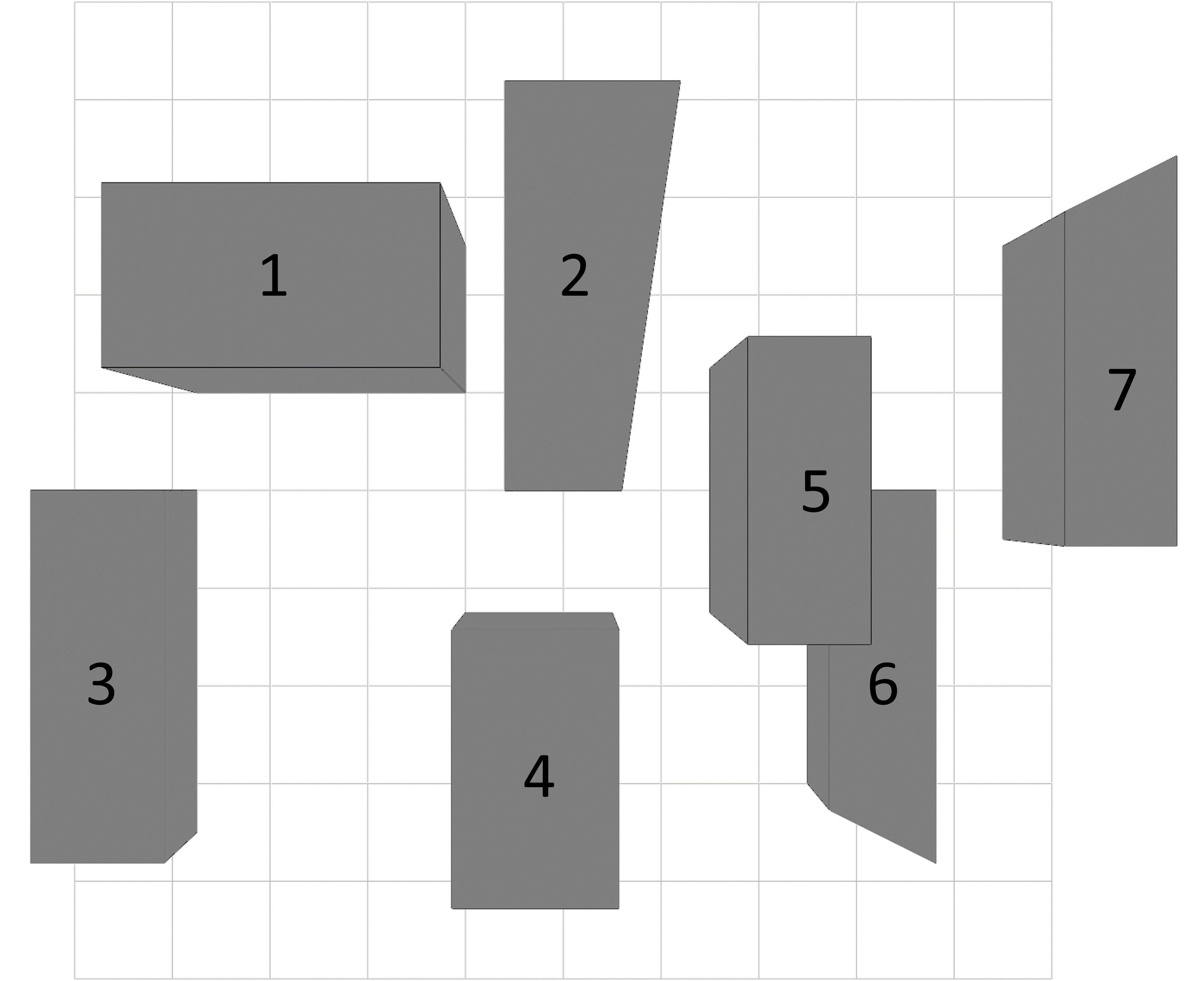}
    \caption{Simulation Scenario: Top Down View of Building Layout and IDs}
    \label{fig:scenario_topdown}
\end{figure}

\begin{figure}[htbp]
    \centering
    \includegraphics[width=\columnwidth]{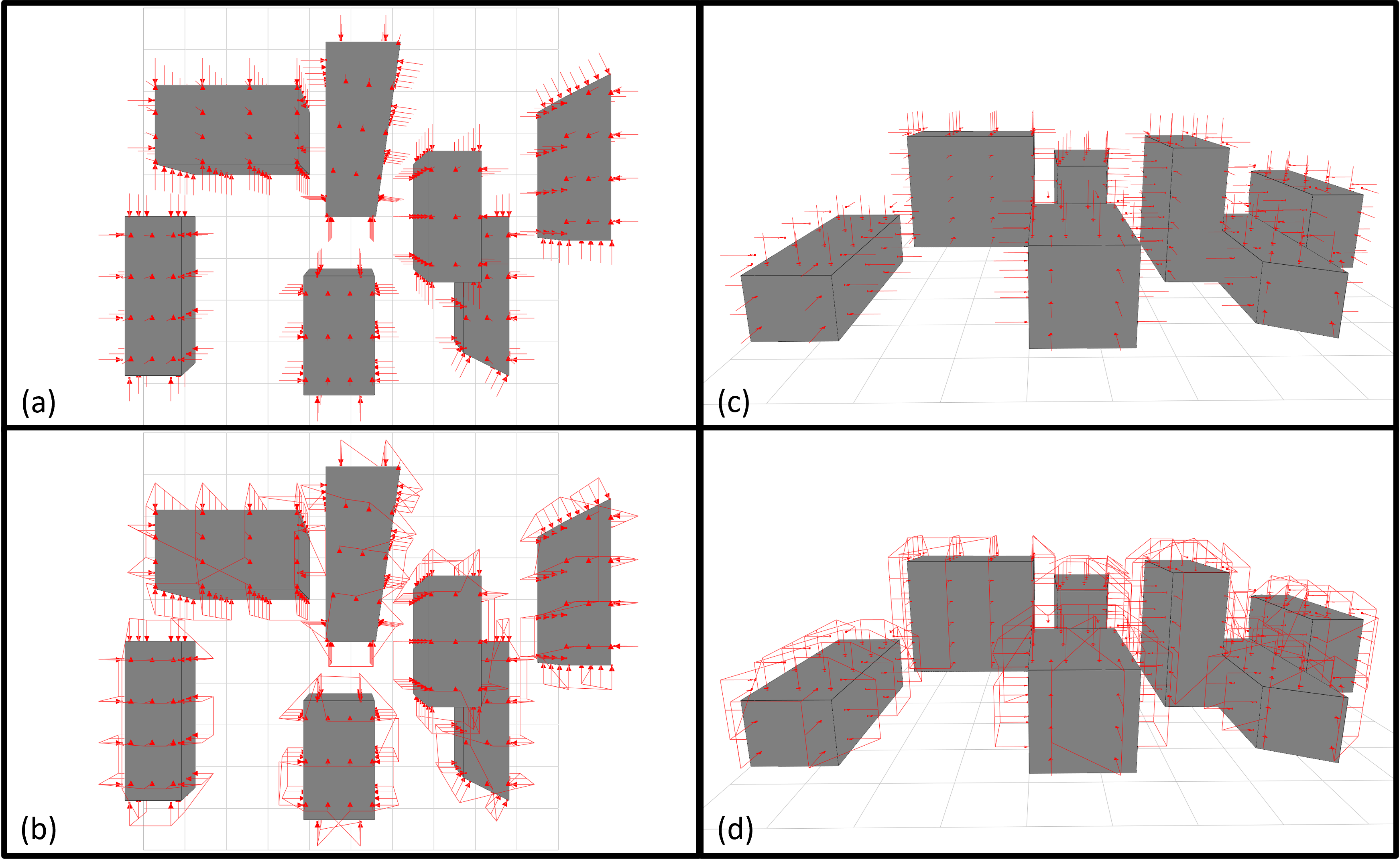}
    \caption{Simulation Scenario: (a) Top down with viewpoints, (b) Top down with closed paths, (c) Side view with viewpoints, (d) Side view with closed paths}
    \label{fig:scenario_viewpoints_tsp}
\end{figure}

We implement and test our proposed algorithm using the Swarm Research Simulator used in our prior work (\cite{leong2021intelligent, leong2022pheromone}). Our scenario consists of 7 buildings ($\lvert \bm{B} \rvert = 7$) modeled as right prisms as shown in Fig~\ref{fig:scenario_topdown}. Viewpoints and paths are generated for each of the buildings as described in Section~\ref{subsec:viewpoint}. Viewpoints that intersect with another building are filtered out. Fig~\ref{fig:scenario_viewpoints_tsp} shows the result of viewpoint generation and closed paths for our scenario.

Given that there are 401 viewpoints, we assign 100 agents to the 7 buildings such that the ratio of agents to viewpoints is roughly 1:4 for each building as shown in Table~\ref{tbl:buildings}.

Each agent moves at 2m/s and stops at each viewpoint for 3 seconds. We assume that this is required to capture images with greater detail and perform object detection of the facade with greater precision. For our camera FOV, we set $c_d = 10, c_w = 84, c_f = 50$, resulting in $f_w \approx 18$ and $f_h \approx 9$.

\begin{table}
\centering
\begin{tabular}{ |c|c|c|c| }
    \hline
    Building ID & Viewpoints & Agents & Path Length (m) \\
    \hline
    1 & 80 & 20 & 955 \\
    2 & 61 & 15 & 802 \\
    3 & 48 & 12 & 662 \\
    4 & 52 & 13 & 680 \\
    5 & 63 & 16 & 792 \\
    6 & 38 &  9 & 533 \\
    7 & 59 & 15 & 803 \\
    \hline
    \hline
    Total & 401 & 100 & 5227 \\
    \hline
\end{tabular}
\caption{Buildings and Task Details}
\label{tbl:buildings}
\end{table}


Our laptop computer system (Intel i7-11850H with 64GB RAM) running our Swarm Research Simulator was able to run the task allocation and patrol algorithms at 1Hz, as well as the UAV model at 10Hz for 100 agents in real time. Running the Christofides algorithm to solve the TSP problem took on the average 3ms for each building, and each agent took less than 2ms to run the proposed patrol algorithm at 1Hz intervals.

\begin{figure}[htbp]
    \centering
    \includegraphics[width=0.75\columnwidth]{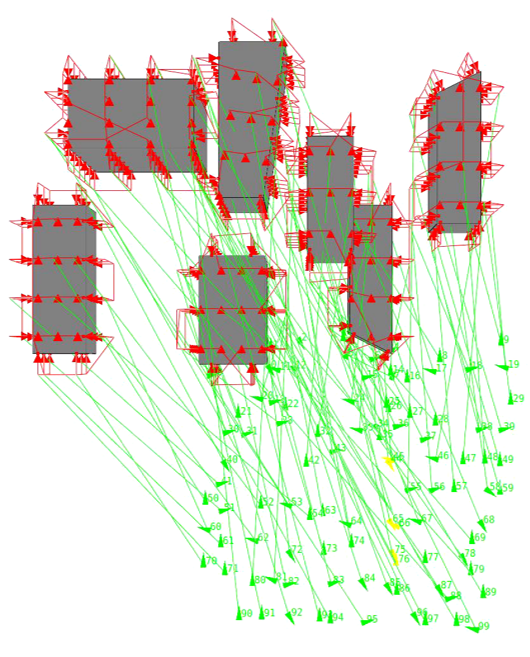}
    \caption{Simulation Scenario: 100 agents performed task allocation and moved towards their assigned buildings.}
    \label{fig:scenario_start}
\end{figure}

\begin{figure}[htbp]
    \centering
    \includegraphics[width=0.75\columnwidth]{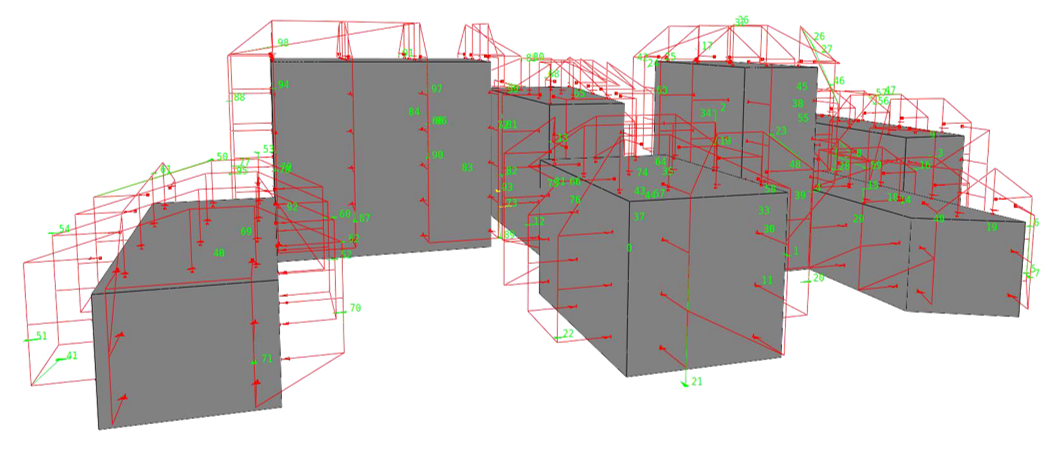}
    \caption{Simulation Scenario: Agents at their assigned buildings execute the patrol strategy to spread out over time.}
    \label{fig:scenario_converged}
\end{figure}

We repeated the scenario 10 times for 1800 seconds each and obtained the following results. 100 agents were initialized and performed task allocation as shown in Fig~\ref{fig:scenario_start}, before moving to their assigned building and executing the patrol strategy, eventually spreading out among all viewpoints as seen in Fig~\ref{fig:scenario_converged}.

\begin{figure}[htbp]
    \centering
    \includegraphics[width=0.75\columnwidth]{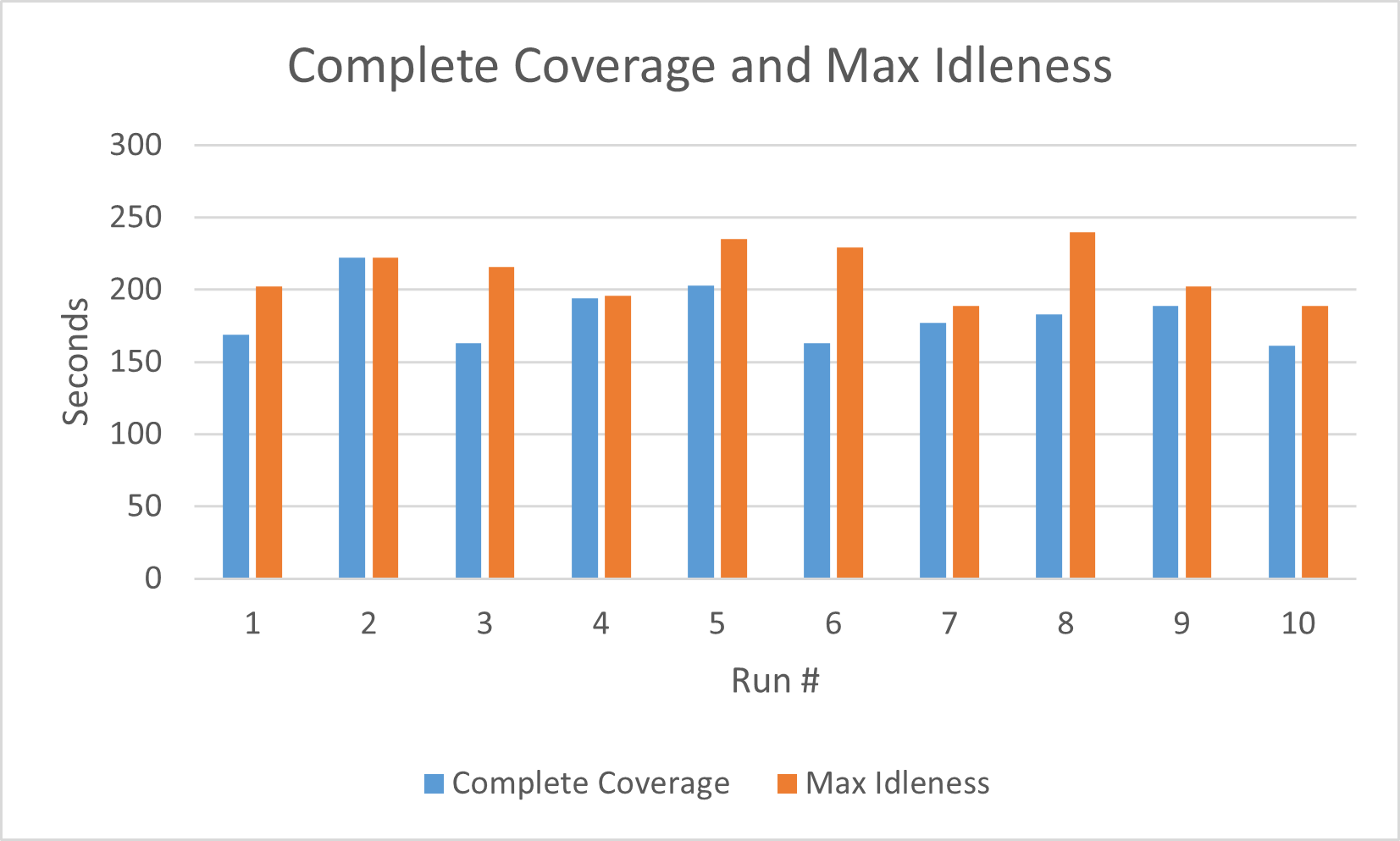}
    \caption{Simulation Scenario: Complete coverage and maximum idleness for 10 experiemental runs show that the patrol strategy is able to converge and achieve complete coverage despite using minimal computation and communications.}
    \label{fig:scenario_complete_coverage_max_idleness}
\end{figure}

We recorded the maximum idleness of all viewpoints at 1Hz for all runs. The idleness for each viewpoint was set to 0 when the scenario started, incrementing over time and resetting to 0 when an agent serviced the viewpoint. This metric allowed us to determine the effectiveness of our proposed method for minimizing the maximum idleness in the scenario and ensuring that the algorithm can converge and provide complete coverage. In addition, we also recorded the time to complete coverage for all runs. This is a measure of how long it takes for every viewpoint to be visited at least once. The results are shown in Fig~\ref{fig:scenario_complete_coverage_max_idleness}

The simulation results show that our proposed approach to minimize the maximum idleness is feasible. Our minimal approach for the patrol strategy is able to constrain the upper bound for the maximum idleness and achieve complete coverage of all viewpoints. The patrol strategy is also scalable since each agent only needs to consider messages and deconflict viewpoints for neighbouring agents.

\addtolength{\textheight}{-6cm}   

\section{Conclusion}
\label{sec:conclusion}
We proposed a decentralized, modular framework and patrol strategy to solve the multi-agent 3D coverage and patrol problem. By breaking the 3D coverage and patrol problem into viewpoint generation, task generation, task allocation, and patrol strategy, we enable each part of the problem to be solved and improved in a modular and incremental fashion.

In our experimental results, we have demonstrated that our proposed patrol strategy can converge to a maximum upper bound for viewpoint idleness and achieve complete coverage, while requiring minimal computation and communications. Our solution achieves the desired results in a decentralized, scalable, and robust manner through the application of simple rules that lead to the desired emergent behavior, albeit at the cost of some optimality. We propose that, in general, swarm algorithms characterized by simple rules that act on local neighborhoods and lead to emergent behavior sacrifice some optimality in favor of achieving scalability.

In future work, we aim to improve the viewpoint generation step by enabling UAVs to utilize heterogeneous cameras, distances, and acute angles to observe surfaces. This will help address challenges such as occlusions or narrow spaces.
                                  
\bibliographystyle{IEEEtran}
\bibliography{3dcoveragepatrol.bib}

\end{document}